%% file: main.tex
\documentclass[conference]{IEEEtran}
\IEEEoverridecommandlockouts
\usepackage{cite}
\usepackage{amsmath,amssymb,amsfonts}
\usepackage{algorithmic}
\usepackage{graphicx}
\usepackage{textcomp}
\usepackage{xcolor}
\usepackage{marvosym}

\usepackage{mdframed}
\mdfsetup{
    linewidth = 1pt, 
    leftline = true, 
    skipabove=\topsep, 
    skipbelow=\topsep, 
}

\usepackage{makecell}
\usepackage{graphicx}
\usepackage{multirow}
\usepackage{svg}
\usepackage{pifont}
\usepackage{enumitem}
\usepackage{tabularx}
\usepackage{bm}
\usepackage{booktabs}
\usepackage{multirow}
\usepackage{subcaption} 
\usepackage{float}  
\usepackage[normalem]{ulem}
\usepackage{url}
\newcommand{\stitle}[1]{\vspace*{0.4em}\noindent{\bf #1\/}}

\def\BibTeX{{\rm B\kern-.05em{\sc i\kern-.025em b}\kern-.08em
    T\kern-.1667em\lower.7ex\hbox{E}\kern-.125emX}}
\begin{document}

\title{How Significant Are the Real Performance Gains? \\ An Unbiased Evaluation Framework for GraphRAG}

\author{
Qiming Zeng$^{{\dagger} {* \thanks{$*$ Equal Contribution}}}$ \quad 
Hao Luo$^{{\dagger}{*}}$ \quad 
Yuhao Lin$^{\dagger}$ \quad
Yicheng Jin$^{\dagger}$ \\
Yuxiang Wang$^{\dagger}$ \quad
Fangcheng Fu$^{\S}$ \quad 
Xiao Yan$^{\dagger}$ \quad 
Jiawei Jiang$^{\dagger,\textrm{\Letter}}$\thanks{$\textrm{\Letter}$ Corresponding author} \\
\vspace*{0.3em}
$^{\dagger}$School of Computer Science, Wuhan University \\
$^{\S}$School of Computer Science, Shang Hai Jiao Tong University \\
\vspace*{0.3em}
$^{\dagger}$ \url{{kirin_z, lohozz, yuhao_lin, jycruina, nai.yxwang,yanxiaosunny,  jiawei.jiang}@whu.edu.cn} \\
$^{\S}$\url{ccchengff@sjtu.edu.cn} \quad 
}

\maketitle

\begin{abstract}
By retrieving contexts from knowledge graphs, graph-based retrieval-augmented generation (GraphRAG) enhances large language models (LLMs) to generate quality answers for user questions. Many GraphRAG methods have been proposed and reported inspiring performance in answer quality. However, we observe that the current answer evaluation framework for GraphRAG has two critical flaws, i.e., \textit{unrelated questions} and \textit{evaluation biases}, which may lead to biased or even wrong conclusions on performance. To tackle the two flaws, we propose an unbiased evaluation framework that uses \textit{graph-text-grounded question generation} to produce questions that are more related to the underlying dataset and an \textit{unbiased evaluation procedure} to eliminate the biases in LLM-based answer assessment. We apply our unbiased framework to evaluate 3 representative GraphRAG methods and find that their performance gains are much more moderate than reported previously. Although our evaluation framework may still have flaws, it calls for scientific evaluations to lay solid foundations for GraphRAG research. 

\end{abstract}


\input{sections/introduction}
\input{sections/relatedwork}

\input{sections/question_part}

\input{sections/evaluation_part}
\input{sections/question_experiment}

\input{sections/experiments}

\input{sections/conclution}
\clearpage
\input{sections/AI}
\bibliographystyle{IEEEtran}
\bibliography{IEEEabrv,main}


\end{document}

%% file: sections/introduction.tex
\section{Introduction}

Large language models (LLMs) have achieved tremendous success in generating answers to user questions~\cite{hadi2023survey,hadi2023large,jiang2021can}. 
However, they may produce wrong or fictitious responses when lacking specific knowledge to answer the questions~\cite{bang-etal-2025-hallulens,huang2023survey,tonmoy2024comprehensive}, e.g., due to outdated training data or missing domain-specific knowledge. 
In these cases, \textit{retrieval-augmented generation} (RAG)~\cite{lewis2020retrieval} improves answer quality by retrieving contexts related to the user question from an external database and feeding these contexts along with the user question to 
the LLMs for answer generation. Among RAG methods, \textit{graph-based RAG} (a.k.a., GraphRAG) is particularly popular, which uses knowledge graphs as the external databases~\cite{procko2024graph,he2024g,wu2025medical}.
This is because knowledge graphs are powerful knowledge representations, where nodes represent real-world entities and edges represent the complex relations between the entities~\cite{chen2020review,hogan2021knowledge,li2024vqft}.

\begin{figure}[t]
    \centering
    \includegraphics[width=0.5\textwidth]{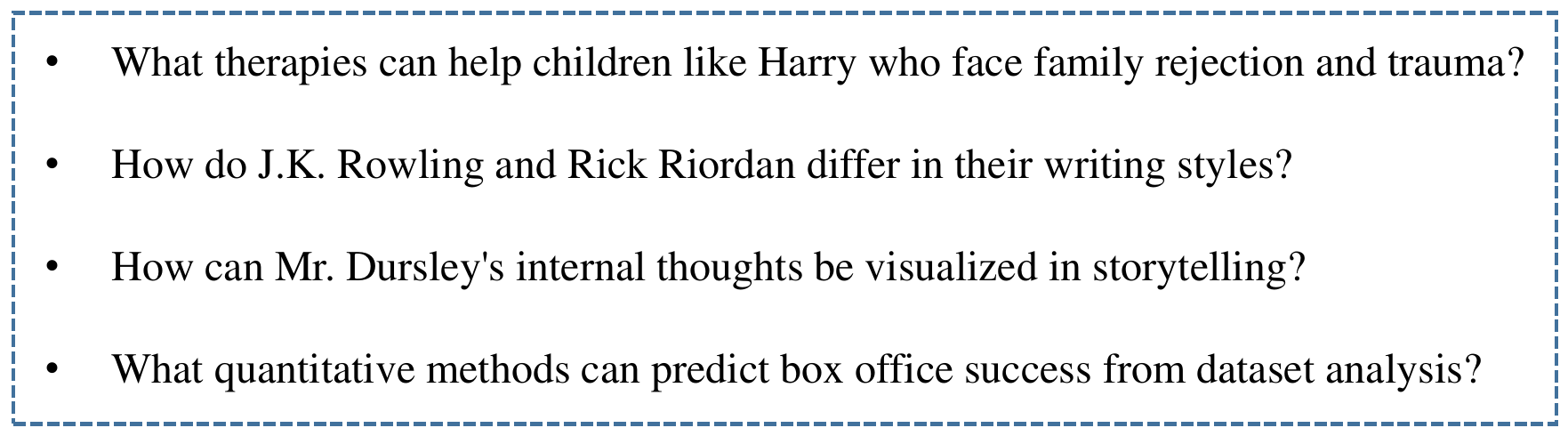}
    \caption{Questions produced for Harry Potter novel by the summary-based method in current evaluation framework.}
    \label{fig:introquestion}
\end{figure}

\begin{figure}[!t]  
    \centering  
        \includegraphics[width=0.7\columnwidth]{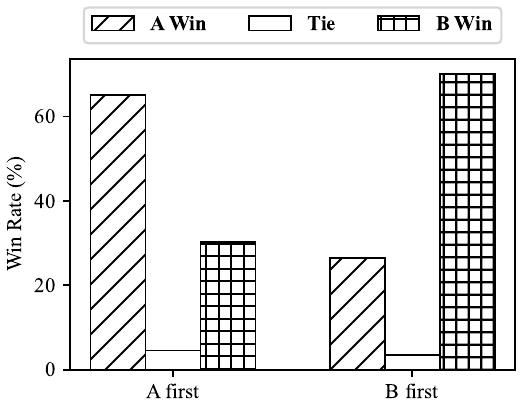}  
    \caption{The effect of position bias. We compare answers generated by LightRAG in 2 runs for the same questions but change the order of the two answers (i.e., `AB' and `BA') when feeding them to the LLM. }  
    \label{fig:positionbias}  
\end{figure}

Many GraphRAG methods have been proposed, and they mainly differ in how to traverse the knowledge graph and retrieve the related contexts. For instance, Microsoft GraphRAG (called \textit{MGRAG} hereafter to avoid confusion with GraphRAG)~\cite{edge2024local} performs community detection on the knowledge graph and generates text summaries for the communities offline, and during online question answering, it selects the community summaries that have large relevance scores for the user question. LightRAG~\cite{guo2024lightrag} first extracts keywords referring to entities and relations from the user question and then retrieves these entities and relations along with their 1-hop neighbors from the knowledge graph. FastGraphRAG (called \textit{FGRAG} hereafter)~\cite{fastgraphrag2024} first extracts entities from the user question, then retrieves relevant contents from the knowledge graph according to semantic similarity w.r.t. the extract entities (measured by their vector embeddings), and finally ranks the contents according to their PageRank scores to select the top ones. GraphCot~\cite{jin2024graph} adopts chain of thought~\cite{wei2022chain} and involves an agent (which is an LLM) to determine the node or edge to check in each graph traversal step. Please refer to~\cite{procko2024graph,han2024retrieval} for a survey of GraphRAG methods. 


Since human inspection is expensive, GraphRAG methods are usually evaluated with LLMs for their answer quality. In particular, an evaluation framework consists of two parts, i.e., \textit{question generation} and  \textit{answer assessment}. For question generation,  MGRAG~\cite{edge2024local} and LightRAG~\cite{guo2024lightrag} extract the head and tail of each passage in the dataset as the passage summary, and the summaries of all passages are aggregated as dataset summaries, which are fed to an LLM for question generation.
For answer assessment, an LLM is requested to compare the answers produced by two RAG methods for the same question, in aspects such as \textit{Comprehensiveness}, \textit{Diversity}, \textit{Empowerment}, and \textit{Directness}, e.g., using a prompt like `\textit{Please choose the better answer from A and B for question Q in term of $\cdots$}', where $A$ and $B$ are the answers produced by two RAG methods. An overall winner is determined based on the separate winners of each aspect, and the win rate of a method is the percentage of questions it wins.



We observe that the current method for GraphRAG evaluation has two critical flaws that hinder scientific performance assessment, namely unrelated questions and evaluation biases.

\stitle{Unrelated questions}. Figure~\ref{fig:introquestion} lists some example questions generated using the \textit{summary-based method} in existing evaluations for the Harry Potter novel. It is obvious that these questions are overly broad and vague. Moreover, they are not (closely) related to the Harry Potter story. This is because the LLM is only provided with vague summaries of the passages in the dataset for question generation, and as a result, the questions do not involve the fine-grained details of the passages. However, GraphRAG is expected to retrieve fine-grained details from knowledge graphs to supplement the LLMs for answer generation, and such ability can not be evaluated using the questions in Figure~\ref{fig:introquestion}.

\stitle{Evaluation biases}. We find that the current evaluation protocol has three biases that hinder scientific performance assessment, i.e., \textit{position bias}, \textit{length bias}, and \textit{trial bias}. In particular, position bias means that simply changing the positions of the two compared answers in the evaluation prompt has a significant impact on the evaluation result. We show such an example in Figure~\ref{fig:positionbias}, where the positions of two answers `\textit{A}' and `\textit{B}' are changed from `\textit{AB}' to `\textit{BA}'. Since we compare the same GraphRAG method (i.e., LightRAG) with itself, the result should be a tie. However, LLMs favor the answer that comes in the front, and the win rates can differ by over 30\%. Similarly, length bias means that LLMs favor longer answers, and trial bias means that an LLM can produce different results when evaluating two answers multiple times.


To tackle the flaws of current GraphRAG evaluation, we design an unbiased evaluation framework with two innovations.

\stitle{Graph-text-grounded question generation}. To ensure that the questions correspond to fine-grained details of the dataset, we use the knowledge graph derived from the dataset to guide question generation. In particular, to generate a question, we first sample a structure from the knowledge graph (i.e., node, edge, or subgraph) and then feed the structure together with its related source contexts to an LLM for question generation. This graph-text-grounded method allows us to configure the type of the generated question, i.e., related to a node, a relation, or a subgraph. We show with concrete examples and a user study that our questions are much more related to the dataset than those generated using the summary-based method.

\begin{figure}[!t]  
	\centering  
	\includegraphics[width=0.8\columnwidth]{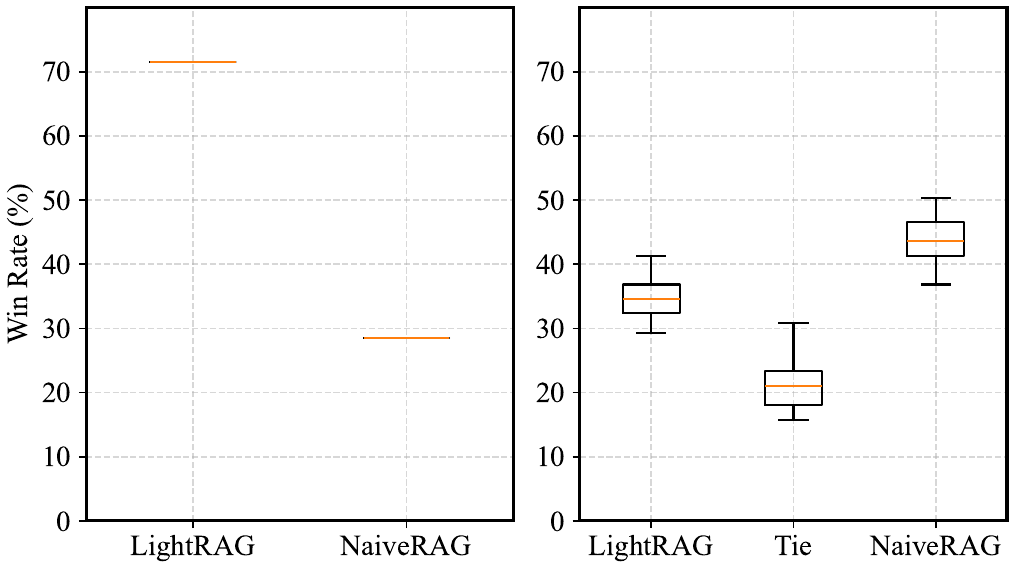} 
	\caption{The evaluation results using the existing method (left) and our unbiased framework (right). Note that existing method does not allow ties. In the box plot (right), the red line is the median, and the lower and upper box boundaries are the 25th and 75th percentiles, respectively.}  
	\label{fig:introbias}  
\end{figure}

\stitle{Unbiased evaluation procedure}. We design a comprehensive evaluation procedure to eliminate the aforementioned biases. In particular, the position bias is handled by evaluating both `\textit{AB}' and `\textit{BA}', and a tie is declared if the overall scores obtained by the two answers are the same. For length bias, we first let two methods generate their answers and then perform answer alignment to align the shorter answer with the longer one in length. For trial bias, we conduct multiple rounds of evaluation with each round covering all the questions and report the statistics of the rounds (e.g., median, percentiles).

We applied our unbiased evaluation framework to compare 3 representative GraphRAG methods, i.e., MGRAG, LightRAG, and FGRAG, and directly retrieving the text chunks without knowledge graph (called NaiveRAG)~\cite{gao2023retrieval} is included as a baseline. The results show that the significant performance gains reported by existing research may be caused by the biases, and after eliminating the biases, the  
performance gains become much more moderate or even vanish. Figure~\ref{fig:introbias} provides such an example, where LightRAG originally outperforms NaiveRAG with a win rate of 72\% vs. 28\% but NaiveRAG slightly outperforms LightRAG after eliminating the biases. This is because LightRAG usually generates longer answers than NaiveRAG, and the original evaluation always places its answer before NaiveRAG. We do observe that FGRAG outperforms the others but again the performance gains are moderate, i.e., below 10\% in win rate expect when compared with LightRAG. 



To summarize, we make the following contributions:

\begin{itemize}

\item We observe that the current method for GraphRAG performance evaluation has critical flaws, i.e., \textit{unrelated questions} and\textit{ evaluation biases}, that may lead to biased or wrong conclusions.

\item We design a new evaluation framework for GraphRAG, which tackles the flaws of the existing method with \textit{graph-text-grounded question generation} and \textit{unbiased evaluation procedure}. 

\item We compare representative GraphRAG methods using our evaluation framework. The results suggest that existing GraphRAG researches are overly optimistic about their performance gains.    

\end{itemize}

We note that our evaluation framework may still be far from perfect. However, given the rapid advances of GraphRAG methods, our work does show that a scientific evaluation framework for their performance is lacking, and without it, we cannot assess whether GraphRAG researches make real progress. We hope that our work can inspire more work on performance evaluation.




%% file: sections/relatedwork.tex


\section{Related Work}

\subsection{GraphRAG Methods}

The Retrieval-Augmented Generation (RAG)~\cite{lewis2020retrieval,gao2023retrieval,siriwardhana2023improving,11113067} system enhances large language models (LLMs) by integrating external knowledge bases, enabling real-time access to up-to-date information without retraining. This approach addresses LLMs’ limitations in handling new or domain-specific knowledge while reducing training costs and mitigating the "hallucination" problem~\cite{ji2023towards,zhang-etal-2025-law,mcdonald2024reducing,wu2025multirag,11113059}, where models generate factually incorrect content. Traditional Naïve RAG achieves this by converting external data into text chunks~\cite{11112897}, vectorizing them, and retrieving the most relevant segments for question answering~\cite{arslan2024survey, fan2024survey,10.1145/3709661}. However, real-world knowledge often exists in graph structures, where entities (nodes) and their relationships (edges) play a key role~\cite{zhang2025survey}. This limitation has driven the evolution from text-based RAG to GraphRAG, which leverages graph-structured knowledge for more accurate and contextually rich retrieval and generation.

MGRAG~\cite{edge2024local} solution first extracts entities and relationships from text chunks, then generates dedicated summaries for these extracted elements as attributes. It then builds a graph-based knowledge structure composed of nodes and edges, runs community discovery algorithms~\cite{rossetti2018community} on this graph to generate community summaries, and finally reorders these summaries as auxiliary context for question answering.

LightRAG~\cite{guo2024lightrag} proposes a dual-level retrieval paradigm, which first extracts entity-related and relationship-related keywords from user queries, then uses these keywords to retrieve corresponding entities and edge relationships in the knowledge graph, and further performs one-hop expansion. Finally, LightRAG integrates and rearranges the node and edge information retrieved from the two hierarchical levels together with there source context to form the final context for answer generation.

FGRAG~\cite{fastgraphrag2024} uses keyword retreival and further introduces a personal PageRank algorithm~\cite{berkhin2005survey} to reorder the retrieved nodes and edges to further enhance the relevance and reliability of the retrieval results. In addition, FGRAG provides a graphical knowledge view that allows users to observe and navigate the retrieval process intuitively, which enhances the interpretability and debuggability of the system.

Other methods include GraphCoT~\cite{jin2024graph}, which leverages LLMs to real-time guide subsequent retrieval information in each retrieval round; ToG~\cite{sun2023think}, which uses LLMs for information filtering and path exploration during retrieval; RoG~\cite{luo2023reasoning}, which employs fine-tuned models to generate retrieval path plans; HippoRAG~\cite{jimenez2024hipporag,gutierrez2025rag}, which proposes retrieving with triples, incorporates passage nodes, and calculates passage node rankings via the PPR algorithm; KET RAG~\cite{huang2025ket}, which divides documents into core and non-core ones, constructing knowledge graphs and Text-Keyword graph respectively for each; and HyperGraphRAG~\cite{luo2025hypergraphrag}, which models edge relationships as hyperedges to accommodate richer relationship types.

\subsection{Current Evaluation Methods for GraphRAG}

GraphRAG evaluation consists of two main components, i.e.,  \textit{question generation} and \textit{answer evaluation}~\cite{yu2024evaluation}. 
Question generation produces a set of user queries to test the GraphRAG methods, while answer evaluation asses the quality of the answers generated by the GraphRAG methods~\cite{salemi2024evaluating}.

For question generation, MGRAG~\cite{edge2024local} and LightRAG~\cite{guo2024lightrag} employ a summary-based method with three steps. 
The first step extracts the beginnings and ends of the articles in the dataset. 
For an article, the beginning part usually outlines the topic and provides background information, while the ending part typically contains the conclusion or summary.
In the second step, the extracted segments are concatenated to form a summary of each article and a dataset description is obtained by stacking the summaries of all articles.
The third step feeds the dataset description to an LLM for question generation, using prompts like `\textit{Given the following description of a dataset, please generate questions for potential user $\cdots$}'.


Answer evaluation is usually conducted by using an LLM as automated judge~\cite{chang2024survey,kocmi2023large,lin2023llm,chiang2023can}, and two answers for the same question are compared each time~\cite{wang2024evaluating,lyu2024crud,salemi2024evaluating,chen2024benchmarking,han2024rag}. In particular, the GraphRAG methods are first instructed to generate answers for a set of questions. Then, the answers produced by two methods for the same questions are paired for comparison. The LLM  is tasked with selecting a superior answer from the pair in some predefined aspects (e.g., \textit{Comprehensiveness}, \textit{Diversity}, \textit{Empowerment}, and \textit{Directness}), and each aspect comes with specific criteria as text instructions. For each aspect, the LLM selects a winner and provides an explanation justifying its decision. Subsequently, the LLM integrates the evaluations across all the aspects to determine an overall winner, e.g., using a prompt like `\textit{select an overall winner based on aspects including$\cdots$}'. For two GraphRAG methods, win rates are computed as the percentages of questions they win in the direct comparison. We note that existing researches place the answers of two GraphRAG methods in a fixed position (e.g., always `AB') in the evaluation prompt.

Some researches~\cite{fastgraphrag2024,jin2024graph} adopt datasets with predefined question-answer pairs. In these cases, answer quality can be assessed using  the similarity between the generated answers and the ground truth answers. 
However, this approach is susceptible to training data contamination~\cite{magar-schwartz-2022-data}, wherein the questions and answers from public datasets may have been included in the LLM's training data, and thus it is difficult to discern whether the generated answers are  attributed to the GraphRAG method or the LLM.
Human experts can provide highly reliable evaluations for answer quality~\cite{wang2023evaluating} but are very expensive to employ. As such, an automated evaluation framework is favorable due to its good generality and low cost.

%% file: sections/question_part.tex
\section{Graph-text-grounded Question Generation}

As shown in Figure~\ref{fig:introquestion}, the questions generated by the current summary-based method are too macroscopic to be answered using the dataset's knowledge or even remain relevant. 
To tackle its defects, we propose a \textit{graph-text-grounded question-generation method} to generate more qualified questions.


\subsection{Question Generation}
Our graph-text-grounded question generation leverages the knowledge graph to produce questions that are closely related to the dataset and mainly contains two steps: \textit{knowledge graph construction} and \textit{graph-text-grounded question generation}.

\stitle{Knowledge graph construction}. We first segment the dataset articles into fine-grained chunks and employ an LLM to extract the entities and relations~\cite{wang-etal-2025-gpt}. 
We prompt the LLM to conduct deeper mining, e.g., using prompts like `\textit{identify the missing entities and add them}'. 
The extracted entities and relations are enriched with the descriptions derived from their source articles. 
These descriptions are stored in a knowledge graph, with the duplicate entities merged to maintain consistency and completeness.
After establishing the knowledge graph, we will associate each entity (node and relationship) in the knowledge graph with its corresponding original text paragraph to ensure that the source of the text paragraph is not lost.

\begin{figure}[t]
    \includegraphics[width=0.5\textwidth]{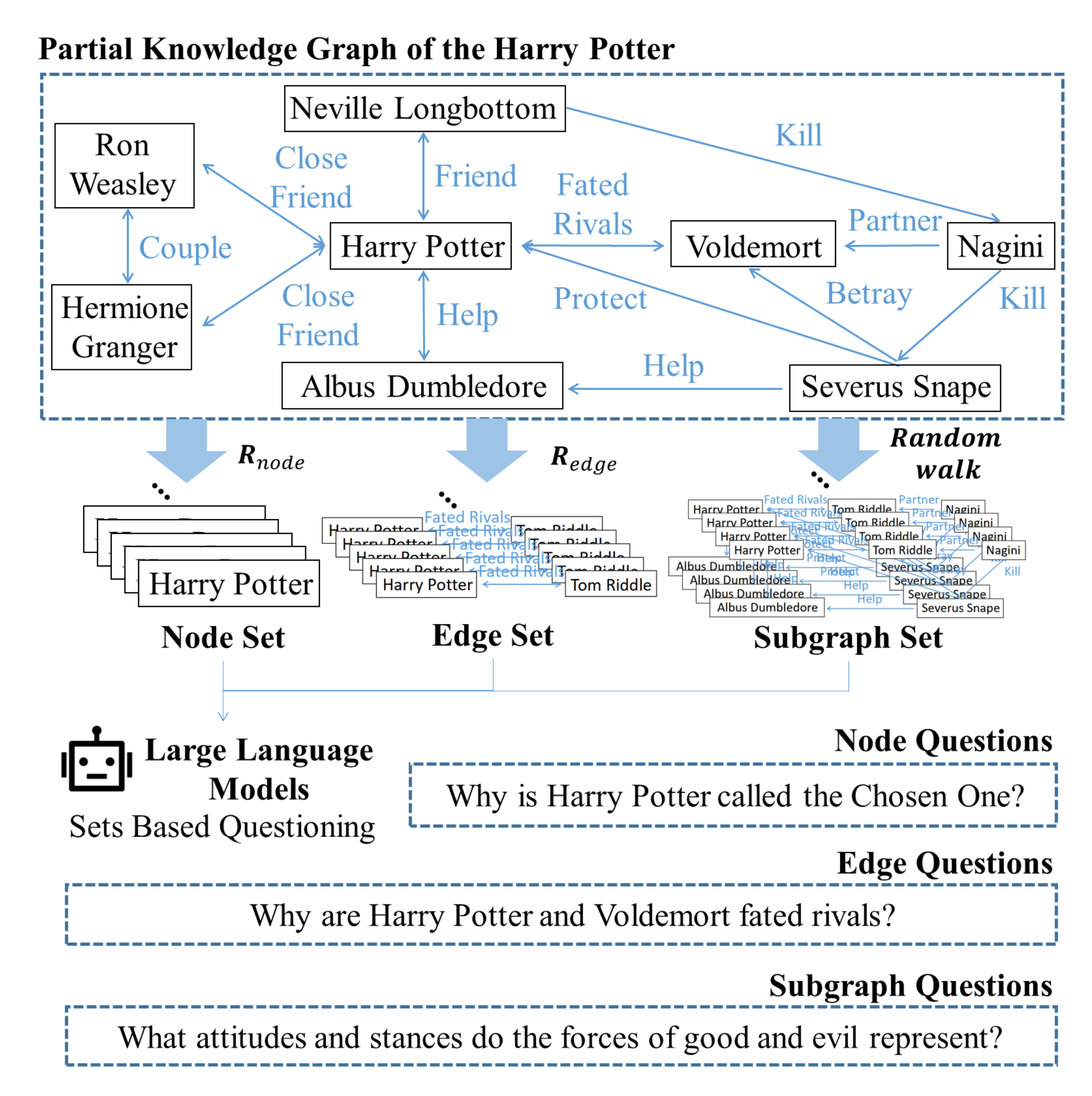}
    \centering
    \caption{Workflow of graph-text-grounded question generation.}
    \label{fig:question}
\end{figure}

\stitle{Question generation principle}.
Current summary-based question generation, relying solely on fixed-length start-end text concatenation, suffers from inaccurate, under-covered summaries, resulting in irrelevant, narrowly-scoped questions. Our graph-text-grounded question generation approach classifies questions into three levels: node-level questions target entity-related information, edge-level questions focus on relation inference knowledge, and subgraph-level questions emphasize the understanding of overall knowledge. Combined with fine-grained sampling, this approach ensures our generated questions are more relevant to the original text.

\stitle{Graph-text-grounded question generation}.
Figure~\ref{fig:question} shows our procedure for graph-text-grounded question generation. We leverage the knowledge graph to generate questions at three levels, i.e.,  \textit{node}, \textit{edge}, and \textit{subgraph}.

At the node and edge levels, we conduct random sampling for the entities and relations of the knowledge graph. These sampled entities and relations form context structures, which are then used to prompt the LLM to generate questions. In particular, at the node level, we instruct the LLM to focus on the contents of the entities to generate relevant questions. At the edge level, we prompt the LLM to focus on the descriptions of the relations to propose inferential questions.
To prevent the quality of the knowledge graph itself from being strongly correlated with the quality of the problem, we will simultaneously take the original text segment corresponding to the entity as supplementary input. In this way, even if the quality of the knowledge graph itself has problems, it can still ensure to the greatest extent that the problem is faithful to the original text.
To generate subgraph-level questions, we first sample a node in the knowledge graph and perform random walk algorithm~\cite{li2015random} from the seed node to extract a subgraph with multiple walk steps. We enforce a minimum size threshold (which is 50 non-repeating hops in default) for the subgraph such that the subgraph encompasses rich entities and relations. If a sampled subgraph falls below the threshold, we resample to ensure adequate size. Given a subgraph, we use it to form context structures that guide the LLM in generating questions that require an overall understanding and reasoning about the subgraph.
For subgraph scenarios, we will first generate summaries of the text chunks corresponding to all entities and relations within the subgraph. Since the text chunks involved in the subgraph at this time are far less than those directly generated for the full text, the summaries will be more accurate. Next, we use the summary as supplementary input to assist in generating subgraph-level questions.


\begin{figure}[t]
    \includegraphics[width=0.5\textwidth]{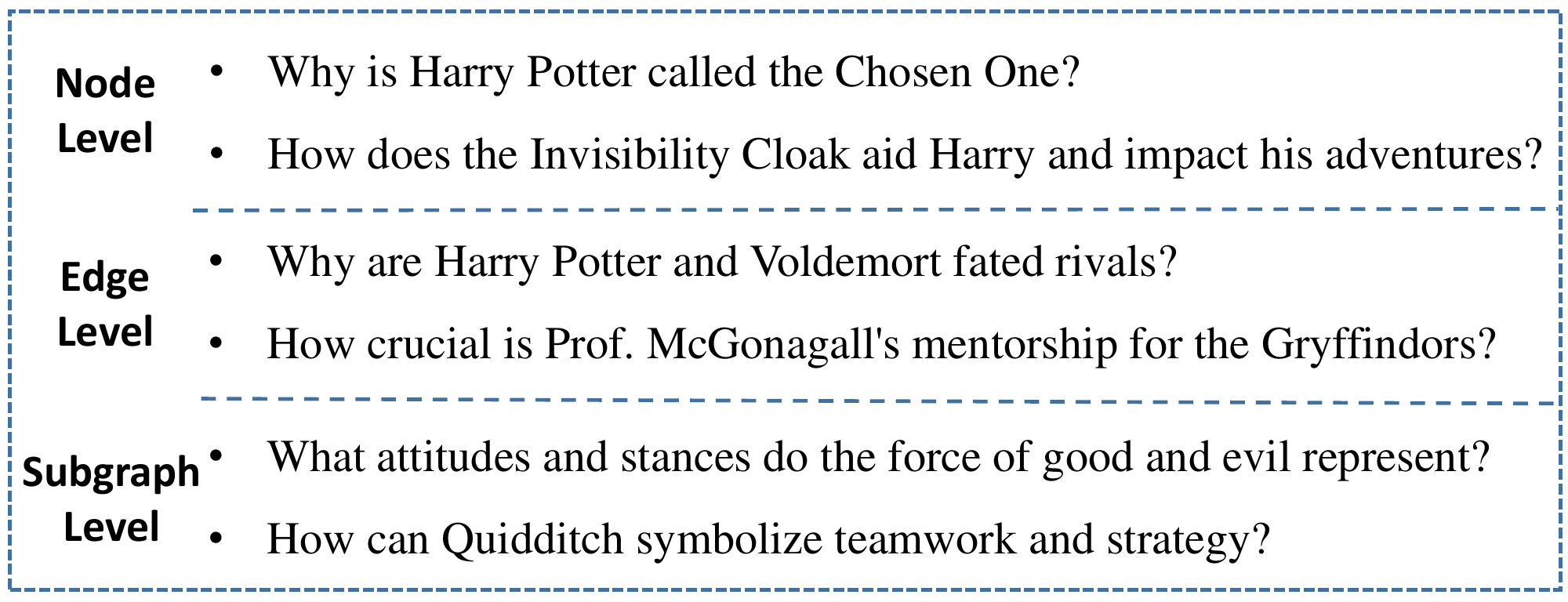}
    \centering
    \caption{Examples of the questions generated by our graph-grounded method for the Harry Potter novel.}
    \label{fig:questionexample}
\end{figure}

\subsection{Question Examples}
We utilize the Harry Potter novel as the dataset to compare the questions generated by the existing summary-based method and our graph-text-grounded method. In particular, Figure \ref{fig:questionexample} shows the questions generated by our method, while Figure~\ref{fig:introquestion} shows the questions generated by the existing method. 
It can be observed that the summary-based method produces questions that are too dispersed and sometimes unanswerable or irrelevant given the underlying dataset. Take the last question in Figure~\ref{fig:introquestion} for example, it is evident that GraphRAG methods cannot forecast the box office receipts using knowledge from the Harry Potter novel. This is because the summary-based method relies on vague summaries of the articles and dataset, which lose the fine-grained details. This problem becomes more significant for larger datasets due to more articles and increased ambiguity.

In contrast, our graph-text-grounded method produces questions that are closely related to multiple levels of the underlying dataset. At the node level, it generates questions focused on the attributes and characteristics of individual entities, testing the ability to retrieve detailed knowledge of specific nodes for GraphRAG methods. At the edge level, it creates inferential questions that explore relations between entities, such as causality or correlation, assessing relational reasoning abilities. At the subgraph level, it formulates macroscopic questions requiring a holistic understanding of multiple interconnected entities, evaluating the ability to synthesize information and comprehend broader structural and semantic contexts. A full evaluation of our proposed graph-text-grounded question generation is further presented in Section~\ref{v}.






%% file: sections/evaluation_part.tex
\section{Unbiased Evaluation Procedure}

In this part, we demonstrate the biases in current evaluation framework and introduce our methods to tackle them. We also present some additional modifications to the evaluation.    

\begin{figure}[t!]  
    \centering  
        \includegraphics[width=0.8\columnwidth]{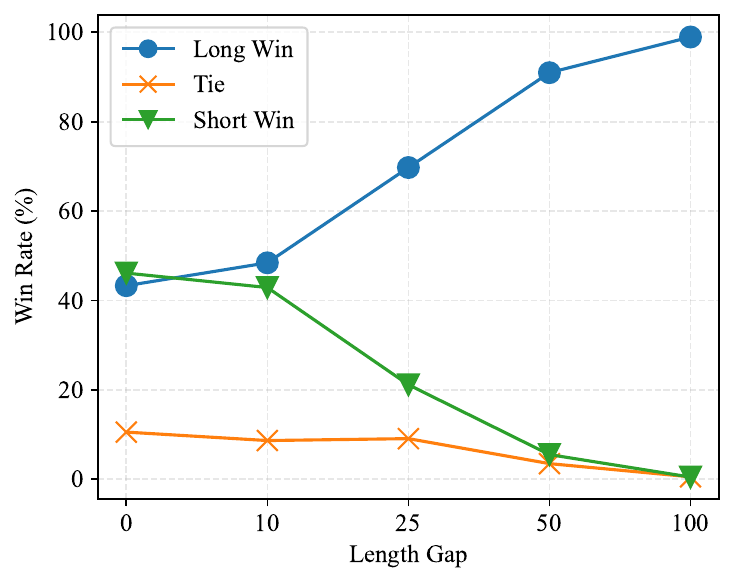}  
    \caption{The effect of length bias. We compare the answers generated by LightRAG from the same questions in two different runs, and the length gap is the token difference between the longer answer and the shorter answer. The average length of the answers is around 200.}  
    \label{fig:lengthbias}  
\end{figure}

\begin{figure}[t!]  
    \centering  
        \includegraphics[width=0.8\columnwidth]{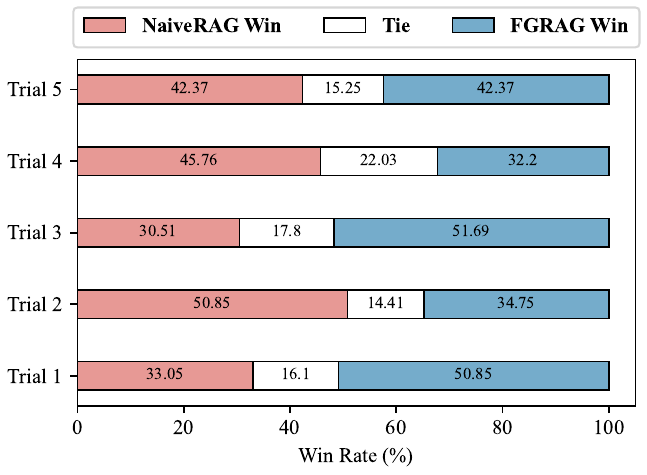}  
    \caption{The effect of trial bias. We compare NaiveRAG and FGRAG on the Agriculture dataset five times with the same set of questions. The win rates differ across the trials.}  
    \label{fig:trialbias}  
\end{figure}

\begin{figure*}[t!]
    \includegraphics[width=1\textwidth]{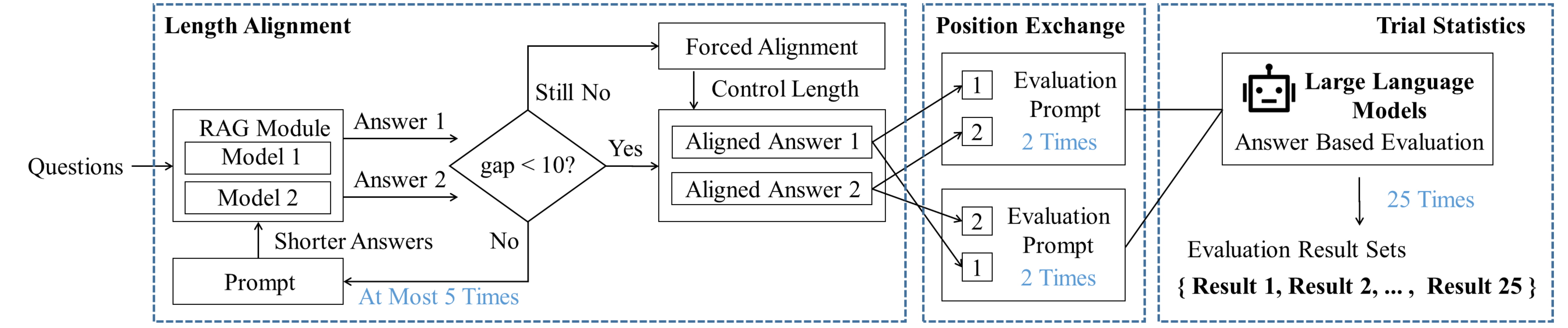}
    \centering
    \caption{An overview of our unbiased evaluation framework.}
    \label{fig:pipline}
\end{figure*}

\subsection{Biases in Existing Evaluation}
Our examination reveals three types of biases when using an LLM as the judge to asses answer quality, i.e., \textit{position bias}, \textit{length bias}, and \textit{trial bias}. These biases severely hinder the fairness of assessment and can lead to absurd conclusions.

\stitle{Position bias.}
As shown in Figure~\ref{fig:introbias}, even when comparing the answers generated by the same GraphRAG method, the method whose answers come in front of the evaluation prompt is much more likely to win. This may be because the attention mechanism of LLMs tends to focus on the initial tokens~\cite{xiao2023efficient}.

\stitle{Length bias.} LLMs tend to perceive the longer answer as having higher quality, Figure \ref{fig:lengthbias} provides an illustration of this phenomenon. For this experiment, we compare answers of different lengths generated by LightRAG. Note that the position bias has already been eliminated using a method that will be presented later. 
Results show that even when the core contents of the answers are similar, a length gap of 25 tokens, which is relatively small compared with the average answer length of 200 tokens, can lead to a win rate gap of over 50\%. This may be because longer answer has more tokens and thus receives more attention. We also observe adding meaningless tokens to an answer may improve its chances of winning.  

\stitle{Trial bias.} LLMs can produce different win rate results when comparing two GraphRAG methods multiple times. We show this phenomenon in Figure~\ref{fig:trialbias}, where NaiveRAG and FGRAG are compared five times using the same set of questions. Note that both position bias and length bias have already been eliminated in this experiment. The results show that different trials lead to contradictory conclusions. In Trial 5, NaiveRAG and FGRAG have a tie; FGRAG wins for Trials 1 and 3, but NaiveRAG wins for Trials 2 and 4. With a single trial, we may arrive at any of the 3 possible conclusions, i.e., tie, NaiveRAG wins, and FGRAG wins. The trial bias is caused by the inherent randomness of LLM generation, which involves sampling the possible tokens.  
  

\subsection{Unbiased Evaluation Framework}

To tackle the evaluation biases discussed earlier, we design an unbiased evaluation framework with three main elements, i.e., \textit{length alignment}, \textit{position exchange}, and \textit{trial statistics}. These three elements correspond to addressing length bias,  position bias, and trial bias respectively. The workflow of our evaluation framework is illustrated in Figure \ref{fig:pipline}.

\stitle{Length alignment.}
To tackle length bias, we align their lengths when comparing two answers for the same question. Given that the GraphRAG methods produce answers of different lengths for the same question, imposing a uniform length constraint may impair their answer quality. As such, we adopt a \textit{generate-adjust} strategy. In particular, both methods first generate answers without restrictions to preserve their logic, and then we adjust the shorter answer to match the length of the longer answer. Instead of directly expanding the shorter answer with the LLM, we adjust the response-generating prompt to include the target length. Compared with expanding the answer, this gives the LLM more room to re-organize the retrieved contexts.

We observe that it is difficult to exactly match the lengths of two answers. Therefore, we set a tolerance threshold for the length difference, which is 10 words by default (observed to have a negligible impact on the evaluation results). If two answers still exceed the threshold after several length adjustments, we apply a forced alignment method. Specifically, we calculate the remaining length discrepancy after the adjustments and instruct the LLM to append tokens to the shorter answer without altering its meaning. Our length aliment procedure can succeed for 85\% of the answer pairs, and we discard the failing answer pairs whose length difference still exceeds the threshold after all adjustments.

{
\renewcommand{\arraystretch}{2}
\begin{table*}
  \caption{The grading criteria for the Directness aspect.}
  \label{tab:show-score}
  \normalsize
  \begin{tabular}{cc}
    \toprule
    Score & Description \\
    \midrule
    \makecell[l]{
    0 point: \\[4pt]
    1 point:\\[4pt]
    2 point:\\[4pt]
    3 point:\\[4pt]
    4 point:\\[4pt]
    5 point:\\} 
    & \makecell[l]{
    The answer is extremely indirect, failing to address the question specifically and clearly.\\[4pt]
    The answer is indirect and deviates significantly from the question, making it hard to discern the intended response.\\[4pt]
    The answer is somewhat indirect, occasionally straying from the question, but still touching on relevant points.\\[4pt]
    The answer is moderately direct, addressing the question with some clarity but could be more specific and focused.\\[4pt]
    The answer is clear and direct, effectively addressing the question with specificity and clarity.\\[4pt]
    The answer is exceptionally direct, precisely and specifically addressing the question without any ambiguity.\\}\\
    \bottomrule
  \end{tabular}
\end{table*}
}

\stitle{Position exchange.}
To address the position bias, we impose position exchange when comparing two answers for the same question. In particular, denote the two answers as $A$ and $B$, position exchange evaluates both $AB$ and $BA$ using two separate prompts. For each position (i.e., $AB$), the LLM is instructed to score the two answers, and the final score of each answer (e.g., $A$) is its average score over the two positions. Position exchange ensures fairness because both answers enumerate the advantageous position (i.e., front) and the disadvantageous position (i.e., back), and thus the average score faithfully indicates answer quality. 

\stitle{Trial statistics.}
To address the trial bias, we compare two answers for the same question multiple times. We adopt two kinds of such repetitions. First, when comparing each position of two answers (i.e., $AB$), we issue the same prompt $N$ times and collect the average scores of these prompts as the scores for a certain position. This is intended to account for the randomness in the LLM's evaluations. 
Second, when comparing two GraphRAG methods, we call it \textit{a trial} when all question-answer pairs are evaluated and obtain win rates. We conduct $M$ such trials to account for the LLM's randomness in generating responses, and thus the evaluation process is re-executed for each trial. By default, we set $N=2$ and $M=25$.

For final evaluation results, we report the statistics of win rates across trials using the box plot~\cite{williamson1989box}, and one example can be found in the right plot of Figure~\ref{fig:introbias}. In particular, for a box plot, the middle line is the median, and the box boundaries are the 25\% and 75\% percentiles. The size of the box (called interquartile range, IQR) reflects the variability of the win rates, and the outliers are depicted as individual points to show extreme deviations. Compared with a single win rate, the box plot provides more information about the results.

A corollary of repeated experiments is the relatively high token overhead. Nevertheless, we argue that a fair and effective evaluation remains necessary and justifiable, even if it entails increased overhead. Furthermore, it is important to note that our evaluation is performed offline, and we adopt a parallelized strategy across different test iterations, which substantially accelerates the overall evaluation process.

\subsection{Enhancement to the Evaluation Protocol}
Besides tackling the evaluation biases, we also enhance the existing evaluation framework with three modifications, i.e., \textit{evaluation perspectives}, \textit{scoring mechanism}, and \textit{tie cases}.

\stitle{Evaluation perspectives.}
We initially adopt the perspectives of MGRAG (i.e.,\textit{Comprehensiveness}, \textit{Diversity}, \textit{Empowerment}, and \textit{Directness}) to evaluate answer quality. However, we observe that diversity is not effective when evaluating a single answer; instead, the relevance of an answer to the question is more indicative for the perceived quality. As such, we supplant the \textit{Diversity} aspect with \textit{Relevance}.

\stitle{Scoring mechanism.}
Existing evaluation instructs the LLM to directly pick a winner based on coarse-grained text comments (for each aspect) that lack quantifiable metrics. To address this limitation, we introduce a scoring mechanism that enhances granularity and objectivity of evaluation. In particular, we ask the LLM to give a score to an answer in each evaluation aspect and provide detailed text instructions to the LLM for scoring. An example of the score grading instructions is provided in Table \ref{tab:show-score}. After scoring all the aspects, we compute the overall score for each answer to determine the overall winner.

\stitle{Tie cases.}
Existing evaluation mandates selecting a winner from the two compared answers, which can be restrictive when the two answers have comparable quality. To address this limitation, we introduce tie cases. Specifically, when the total scores of the two answers are the same, we declare a tie, acknowledging that the answers have similar quality.

%% file: sections/question_experiment.tex
\section{Graph-Text-Grounded Questions Evaluation}
\label{v}

In this section, we conduct a detailed evaluation of questions generated by our proposed graph-text-grounded question generation method. We generate 150 graph-text-grounded questions and 150 summary-based questions respectively. First, we compare the entity distribution of each of the two question sets. Next, we use an LLM to evaluate the relevance between retrieved knowledge and questions, and assign scores accordingly. Finally, we carry out an exhaustive user study. Experiments demonstrate that our graph-text-grounded question generation method achieves superior performance.

\subsection{Question Entity Distribution}

\begin{figure}
    \centering
    \includegraphics[width=1\linewidth]{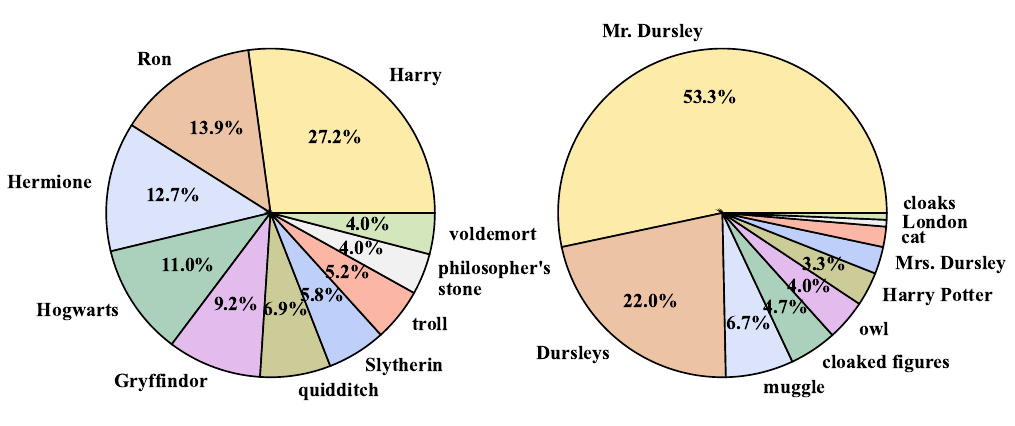}
    \caption{Entity frequency pie chart of graph-text-grounded questions (left) and summary-based questions (right).}
    \label{fig:pie}
\end{figure}

Summary-based question generation methods simply concatenate text segments of a specified length from the start-end of a document to form a summary~\cite{guo2024lightrag}, based on which questions are generated. We argue this approach may limit the LLM’s perspective for question generation, to these short contextual segments, narrowing its focus. Consequently, the generated questions may only cover a small portion of the entities in the dataset while neglecting more content.

We performed entity extraction and statistical analysis on the question sets generated by the two methods. The specific process and results are as follows: first, we prompted an LLM to extract entity information related to the test dataset (i.e., the novel Harry Potter and the Philosopher's Stone) from the given questions; then, we calculated the word frequencies of these extracted entities for each question generation method, and plotted the frequencies of the top-10 entities in Figure~\ref{fig:pie}.

As illustrated in these pie charts, the entity distribution of questions generated by the summary-based method exhibits significant imbalance. Specifically, a substantial proportion (53.3\%) of its questions concentrate on minor characters that appear in the initial segment of the dataset, with Mr. Dursley serving as a representative example.
In contrast, the questions generated by our graph-text-grounded method demonstrate a more uniform entity distribution while covering a broader range of entities. Quantitatively, only 11 distinct entities were extracted from the summary-based questions, whereas 64 distinct entities were identified from the questions generated by our graph-text-grounded method. Furthermore, our method shows a relative focus on the three core characters of the Harry Potter novel, namely Harry, Ron, and Hermione.

This discrepancy stems from the fundamental difference in the two methods’ design: our method samples on the knowledge graph during question generation, which drives the generated questions to cover sampled nodes, edges, and subgraphs. The summary-based method, however, only focuses on a limited range of content explaining why its questions cluster around a small set of entities (like Mr. Dursley) while ours achieve broader, more core-focused entity coverage.

    
    

\subsection{Question-Retrieval Relevance}

\begin{figure}[!t]  
    \centering  
        \includegraphics[width=0.85\columnwidth]{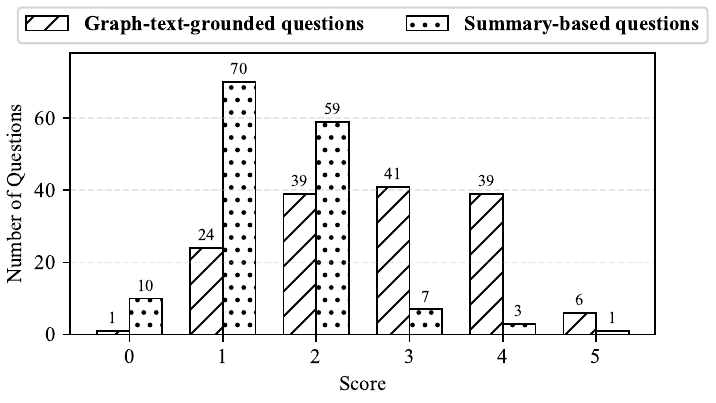}  
    \caption{Relevance score assigned by LLM between retrieval contexts and questions. We used FGRAG to retrieve contexts and instructed the LLM to assign a relevance score (ranging from 0 to 5) for each pair of retrieved context and question.}  
    \label{fig:llmeval}  
\end{figure}

Another limitation of summary-based questions is their inherent information insufficiency, which may hinder identifying highly relevant dataset content. To address this, we designed a supplementary experiment: first, we used FGRAG (a method exhibiting the best performance during subsequent evaluations) to retrieve contextual information for both summary-based questions and graph-text-grounded questions, then paired these contexts with their questions to form "question-retrieval context" pairs. Next, we asked an LLM (i.e., gpt-4o-mini) to rate the relevance of each pair on a 6-point scale (0–5), 0 means completely irrelevant, while 5 means the context suffices to fully answer the question. Results are shown in Figure~\ref{fig:llmeval}.

As clearly shown in the figure, summary-based questions consistently have lower relevance scores, with their distribution heavily concentrated in the 0–2 point range, scores of 0 and 1 alone account for over 50\% of their total ratings.
In sharp contrast, our Graph-text-grounded questions not only have a significantly higher overall score profile but also a more balanced distribution. This balance indicates our questions are more effective at assessing the actual performance of GraphRAG systems. Notably, if even the best-performing model struggles to retrieve question-relevant information, the question itself may be inherently unanswerable; we further explore this issue in the user study section.

\subsection{User Study}

We conducted a user study to evaluate the linguistic quality and content relevance of graph-text-grounded questions, aiming to verify whether they outperform summary-based ones. We provide a questionnaire example in the uploaded materials.

We recruited 40 participants from various academic institutions, aged 20–28. All participants had basic English reading proficiency and a general understanding of the test dataset’s story content (i.e., Harry Potter and the Philosopher’s Stone).

We selected the full text of \textit{Harry Potter and the Philosopher’s Stone} as the test corpus, and used the two methods to generate two different question sets (150 questions each). For evaluation convenience, we randomly sampled 15 questions from the summary-based questions set, and 5 questions from each level of our graph-text-grounded questions set (i.e., \textit{node-level, edge-level, subgraph-level}). These 30 questions in total were randomly ordered and assigned to each questionnaire.

Specifically, we focused on three key evaluation dimensions: \textit{Comprehensibility}, \textit{Specificity}, and \textit{Answerability}. Participants were instructed to evaluate each question along these three dimensions using a five-point Likert scale, where a higher score indicated better quality. The definitions are as follows:

\begin{itemize}

\item \textit{Comprehensibility}: Measures question grammar and clarity. Low scores reflect severe grammatical errors or ambiguity; high scores indicate fluent, well-structured, and easy-to-understand questions.

\item \textit{Specificity}: Measures how clearly a question defines its scope, target, and requirements. Low scores means vagueness or missing key elements; high score indicates clear boundaries and precise wording.

\item \textit{Answerability}: Measures whether a question can be fully answered using only the source text. Low scores mean insufficient source support or reliance on external knowledge; high scores indicate the question can be directly and sufficiently answered from the source text.

\end{itemize}

\begin{figure}[t!]  
    \centering  
        \includegraphics[width=0.8\columnwidth]{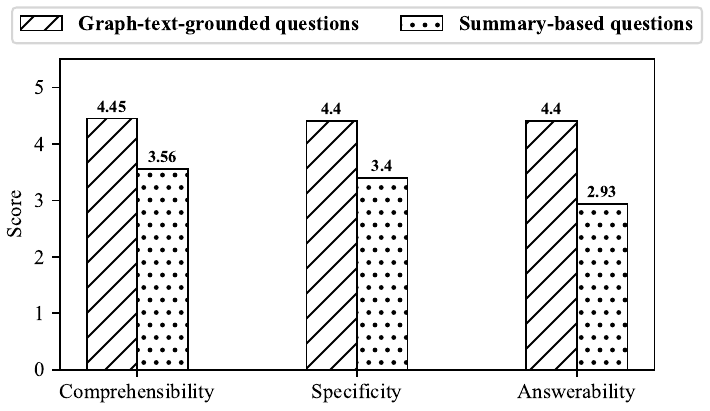}  
    \caption{Results of our user study. We present the average scores given by participants for three dimensions: Comprehensibility, Specificity, and Answerability across both graph-text-based questions(left)and summary-based questions (right). }  
    \label{fig:task-result}  
\end{figure}

The rationale for choosing these three dimensions is that they respectively represent linguistic clarity, information focus, and semantic relevance which are consistent with the widely adopted evaluation standards in prior studies~\cite{2020Exploring,kurdi2020systematic,2024QGEval}.

\begin{figure*}[t] 
        \centering
        \includegraphics[width=0.95\textwidth]{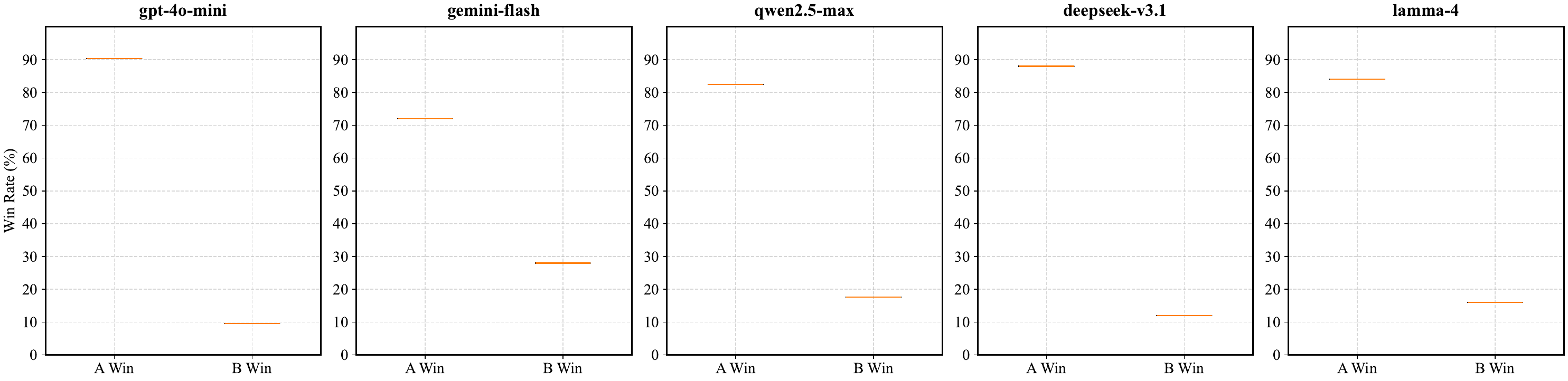}
        \caption{Current framework results. We compare the answers generated by the same GraphRAG method (i.e., LightRAG) using current evaluation method's across different LLMs (i.e., gpt-4o-mini, gemini-flash, qwen2.5-max, deepseek-v3.1, lamma-4). Notably, the method produced the same erroneous conclusion (i.e., LightRAG outperforms itself) for all five LLMs; this universal mistake confirms the bias in the current evaluation framework is broadly applicable and not limited to specific models.}
        \label{fig:crossmodel5}
\end{figure*}

\begin{figure}
    \centering 
    \includegraphics[width=0.4\textwidth]{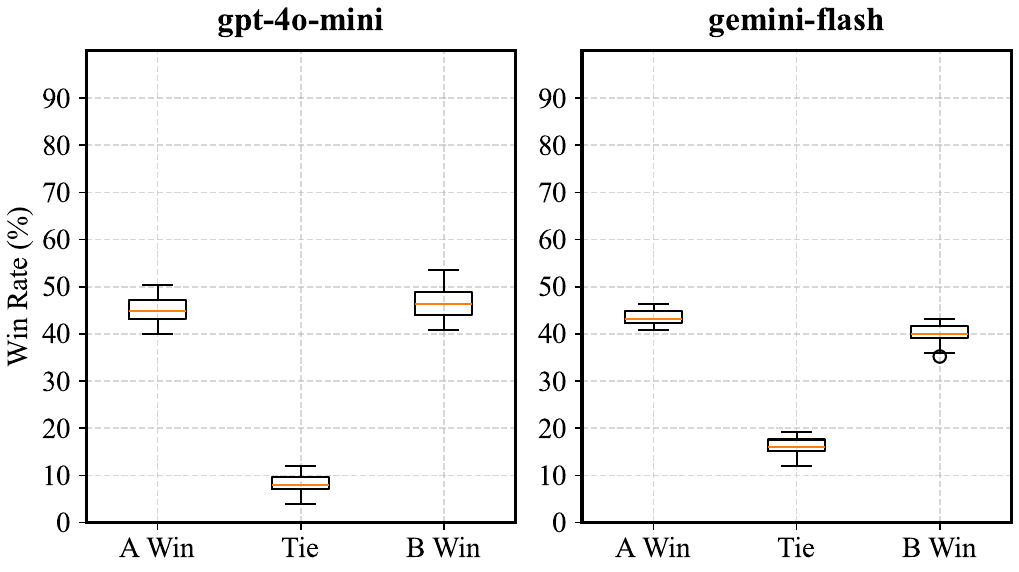}    
    \caption{Sanity check. We compare the answers generated by the same GraphRAG method (i.e., LightRAG) using our proposed unbiased evaluation framework on two different LLM(i.e., gpt-4o-mini, gemini-flash). The result shows our evaluation method correctly determines that the method matches itself, and this finding extends to other LLMs as well.} 
    \label{fig:my senity check} 
\end{figure}

Before formal evaluation, participants got a brief study intro and a summary of the testing corpus (main plot and characters) to help them recall the story. Each participant then independently rated 30 questions across the three dimensions. Responses were collected for statistical analysis and comparing the two question generation methods.

Figure~\ref{fig:task-result} illustrates the comparison of average scores between graph-text-grounded questions and summary-based questions across evaluation dimensions. We make the following observation:
Overall, our graph-text-grounded questions outperformed the summary-based questions in all dimensions, demonstrating the effectiveness of the multi-level sampling strategy in improving the quality of generated questions.

Specifically, our graph-text-grounded questions outperformed summary-based ones in both dimensions: \textit{Comprehensibility} (4.45 vs. 3.56, +~0.89) and \textit{Specificity} (4.40 vs. 3.40, +1.00).
This is because our approach uses multi-level sampling, which better preserves context for more fluent, understandable questions. It also focuses on targeted text segments and added descriptions for clearer scope and intent, avoiding the vagueness of the summary-based question generation method’s summary-concatenation process.

The greatest improvement was in \textit{Answerability}. Our graph-text-grounded questions averaged 4.40, vs. 2.92 for summary-based ones. This shows our questions align more closely with the source knowledge, allowing them to be answered using the text content. In contrast, summary-based questions, which rely on concatenated segment summaries, often lack key context or have semantic gaps, making many unanswerable.



%% file: sections/experiments.tex
\section{Evaluation on Current GraphRAG Methods}


\begin{figure*}[t!]  
    \centering  
    \includegraphics[width=1\textwidth]{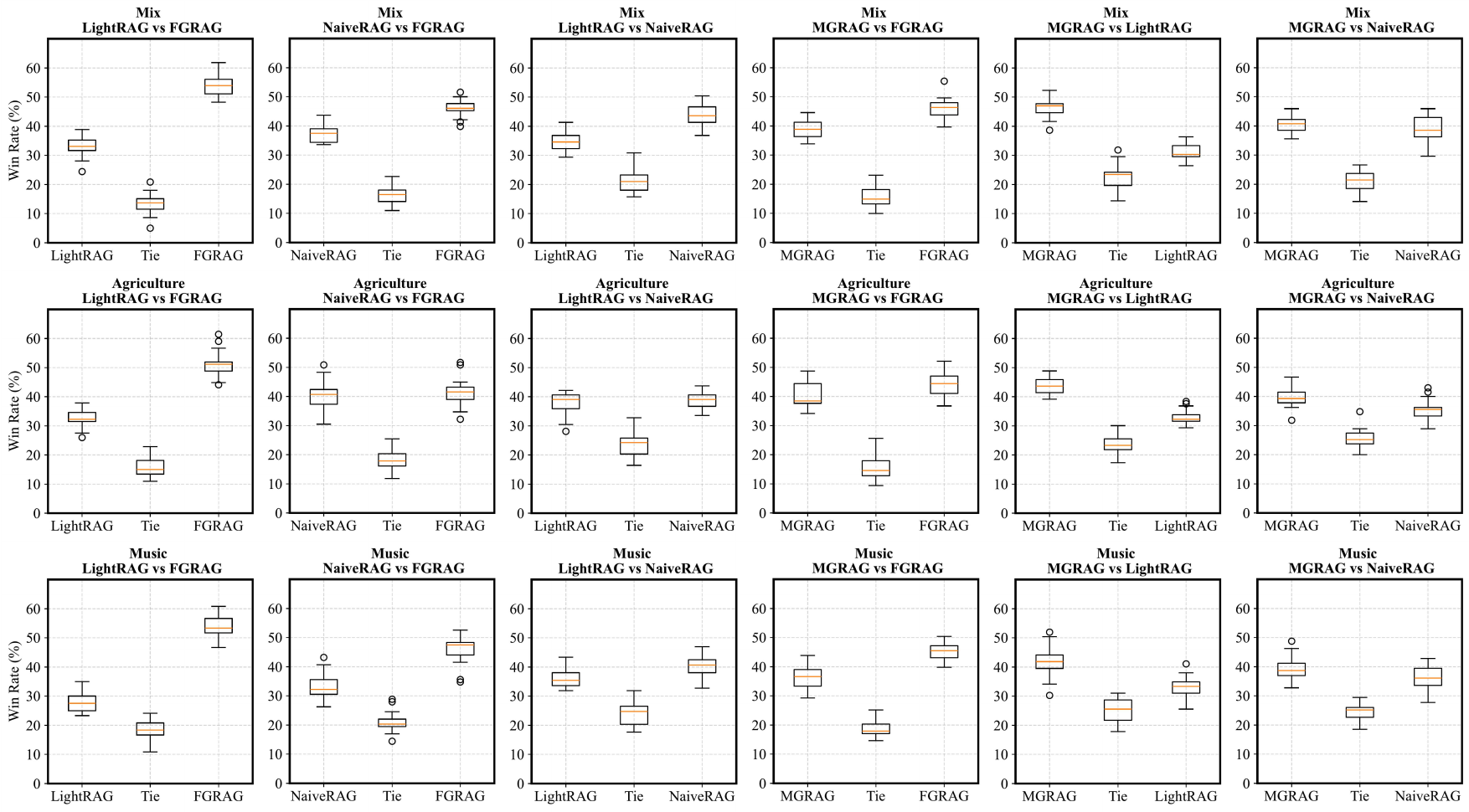}  
    \vspace{-0.8em}
    \caption{One-to-one comparison between the 3 GraphRAG methods and NaiveRAG, each row corresponds to a dataset.}  
    \label{fig:overall}  
\end{figure*}

In this section, we first conduct a sanity check. We then apply our evaluation framework to examine three representative GraphRAG methods, demonstrating that it leads to conclusions that are radically different from those reported previously.

\subsection{Experiment Settings}
\stitle{GraphRAG methods.} We evaluate 3 representative GraphRAG methods, i.e., MGRAG~\cite{edge2024local}, LightRAG~\cite{guo2024lightrag}, and FGRAG~\cite{fastgraphrag2024}, which are popular recently. We include NaiveRAG as a baseline, which directly retrieves text chunks according to embedding similarity w.r.t. the question and does not utilize knowledge graph. We use their publicly accessible codebases to conduct answer generation.


\stitle{Datasets.}
We use the Ultra Domain Benchmark (UDB) datasets ~\cite{qian2024memorag} as the knowledge base. In particular, UDB is a diverse and representative collection compiled from 428 university textbooks and spanning 18 subjects. Each sub-corpus contains between 600,000 and 5 million tokens. We use the following 3 sub-corpus.

\begin{itemize}
    \item {\textit{Mix:}}
    A compact and diverse sub-corpus of UDB. It is the smallest among our three datasets with 5.74 MB.
    \item {\textit{Agriculture:}} It focuses on agricultural contents and covers topics like beekeeping and farming. Its size is 70 MB.

    \item {\textit{Music:}}
    It encompasses a variety of subjects related to music (such as musical styles, biographies, etc). It is the largest among the three datasets, totaling 143 MB.
\end{itemize}

 These datasets are also used in the experiments of LightRAG~\cite{guo2024lightrag}. Additionally, we utilize the \textit{GPT-4o-mini} model as the LLM to generate answers as well as serve as the evaluation judge. We generate 50 distinct questions for each of the 3 levels, i.e., node, edge, and subgraph, using our graph-text-grounded question generation method. As such, the total number of questions is 150.

\stitle{Environments.}
We use a single A6000 GPU with 48 GB of memory on a server equipped with one Intel Xeon Platinum 8375C CPU, and 256 GB of memory.

\subsection{Sanity Check}

Before reporting the main results, we first verify that our proposed unbiased evaluation framework is truly unbiased.

\stitle{Principle.}
We use the same model (i.e., LightRAG) to generate two sets of answers for the same question set. We then conduct evaluations using both the current framework and our proposed framework. An unbiased evaluation should yield substantially consistent win rates for the two sets of answers.

\stitle{Results.}
We conducted tests on several representative LLMs and present result in Figure~\ref{fig:crossmodel5}, we also implement our evaluation framework and show the results in Figure~\ref{fig:my senity check}.
Intuitively, a reasonable evaluation should indicate that the two instances are comparable in answer quality, given that they are essentially the same GraphRAG method. However, current evaluation results show that for gpt-4o-mini, instance $A$ achieves a win rate of 90\%, while instance $B$ only reaches a win rate of 10\%, similar results are observed across other tested LLMs. This translates to an 80\% advantage for $A$ and falsely suggests significant performance gains. Such misleading outcomes stem from two key issues: the answers of $A$ are longer(i.e., length bias) and are placed in the front (i.e., position bias), while the evaluation relies on only a single trial (i.e., trial bias). In contrast, our proposed unbiased evaluation framework correctly identifies that the two instances are essentially on par with each other in terms of overall performance.

\subsection{Results and Analysis}

\stitle{Overall performance.}
We perform a one-by-one comparison for the 3 GraphRAG methods and NaiveRAG and report the results in Figure~\ref{fig:overall}. We can make the following observations.

First, the win rates are generally smaller than reported by previous work, and tie rates are notable, making the performance gaps between the methods small. For instance, LightRAG~\cite{guo2024lightrag} reports that it achieves win rates of 66.70\% (vs. NaiveRAG) and 56.38\% (vs. MGRAG) on the Agriculture dataset, while in Figure~\ref{fig:overall}, the win rates of LightRAG become 39.06\% (vs. NaiveRAG) and 32.33\% (vs. MGRAG) on the same dataset. Moreover, tie rates are over 20\% in most cases. These results are because our evaluation framework resolves the biases and allows ties. Moreover, our evaluation also changes the relative performance of the methods. For example, LightRAG is reported to outperform NaiveRAG~\cite{guo2024lightrag} but our results show that NaiveRAG performs better than LightRAG.

Second, FGRAG performs the best among the methods, followed by MGRAG, then NaiveRAG, and finally LightRAG. This is because FGRAG assigns more concise and informative contexts to entities and relations in the knowledge graphs and adopts the PageRank scores to rank retrieved entities and relations. Microsoft also shows that FGRAG can accurately retrieve information relevant to the question~\cite{fastgraphrag2024}. The results suggest that a high-quality knowledge graph and proper re-ranking for retrieved contents are beneficial for performance.

Third, the performance advantages of GraphRAG methods over NaiveRAG generally improve with dataset size. This is because GraphRAG methods use knowledge graphs for context retrieval while NaiveRAG directly retrieves text chunks. The knowledge graphs help to accurately retrieve information and enable large-span retrieval, which is crucial for large datasets.

\begin{figure*}[t!]  
    \centering  
    \includegraphics[width=1\textwidth]{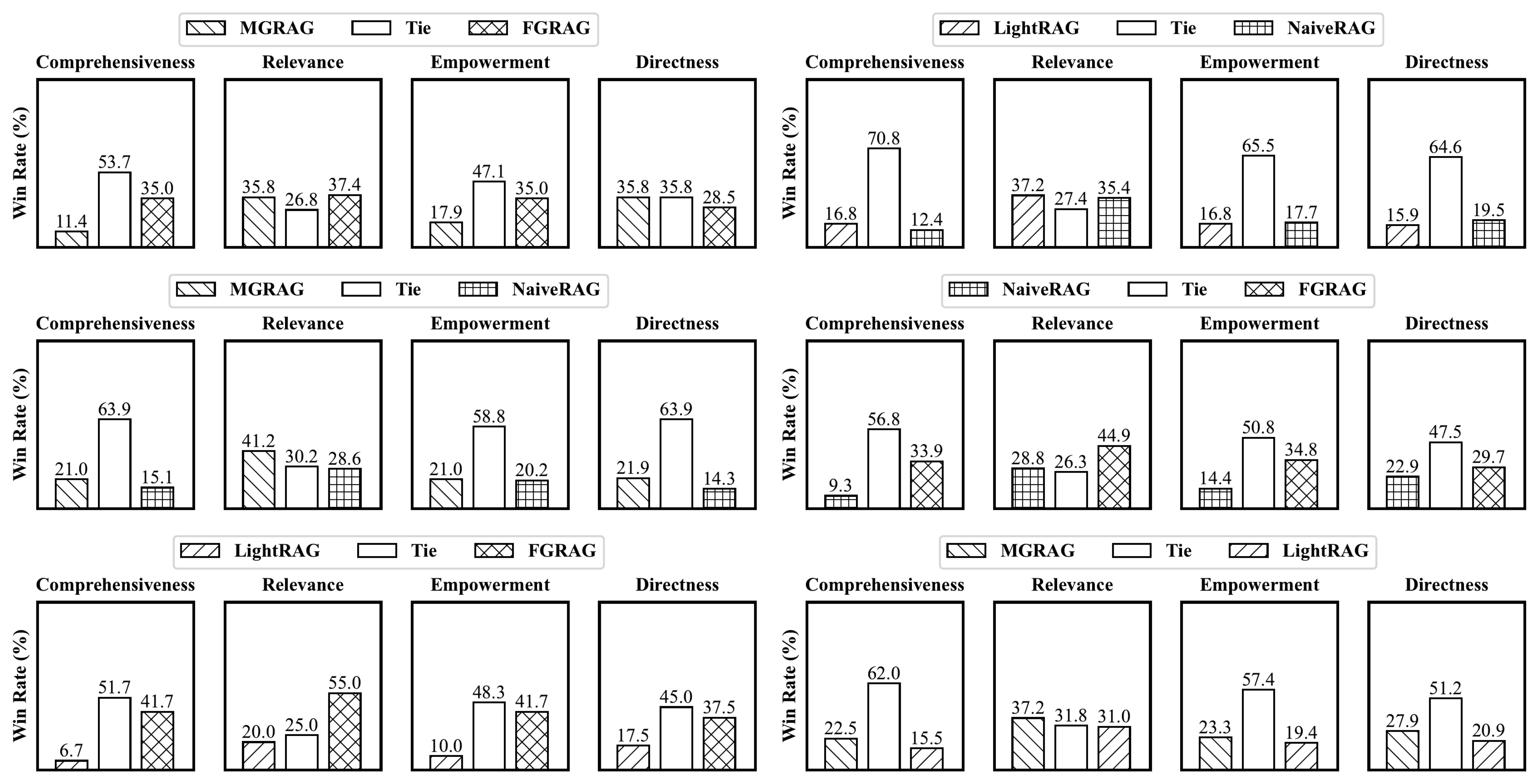}  
    \caption{Detailed win rate comparison for the 4 aspects of answer quality, the dataset is Music.}  
    \label{fig:detail}  
\end{figure*}

\begin{figure}[t!]  
    \centering  
        \includegraphics[width=0.7\columnwidth]{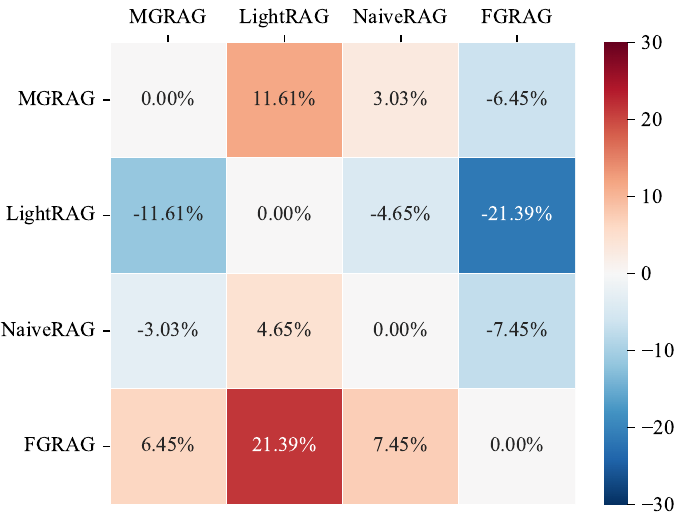}  
    \caption{Relative win rates (defined as row minus column) between the 3 GraphRAG methods and NaiveRAG. A positive value means that the Row method outperforms the Column method. The results are averaged over 3 the datasets.}  
    \label{fig:heatmap}  
\end{figure}

\stitle{Relative win rate.}
To better visualize the performance differences between the methods, we introduce \textit{relative win rate}, which is defined as follows:
\begin{equation*}
\text{Relative Win Rate} = \frac{A_{\text{win}} - B_{\text{win}}}{A_{\text{win}} + B_{\text{win}} + \text{Tie}},
\end{equation*}
where $A_{\text{win}}$ is the number of questions method $A$ wins, and similarly for $B_{\text{win}}$ and $\text{Tie}$.
Relative win rate measures the advantage of one method (i.e., $A$) over another (i.e., $B$). 

Figure~\ref{fig:heatmap} plots the relative win rate between the methods. We observe again that the performance gaps between different GraphRAG methods are moderate. Specifically, with the exception of FGRAG and MGRAG, which improve LightRAG by 21\% and 11\%, respectively, the relative win rates between the other methods are all below 8\%, which is consistent with the results from our overall performance evaluation.

\begin{mdframed}
\em \textbf{Findings:} Many models’ actual improvement is less significant than self-reported. High-quality ranking of retrieved results remains a key component in GraphRAG. GraphRAG becomes more effective as datasets scale up. 
\end{mdframed}



\begin{figure*}[t] 
    \begin{subfigure}[t]{0.3\textwidth} 
        \centering
        \includegraphics[width=\linewidth]{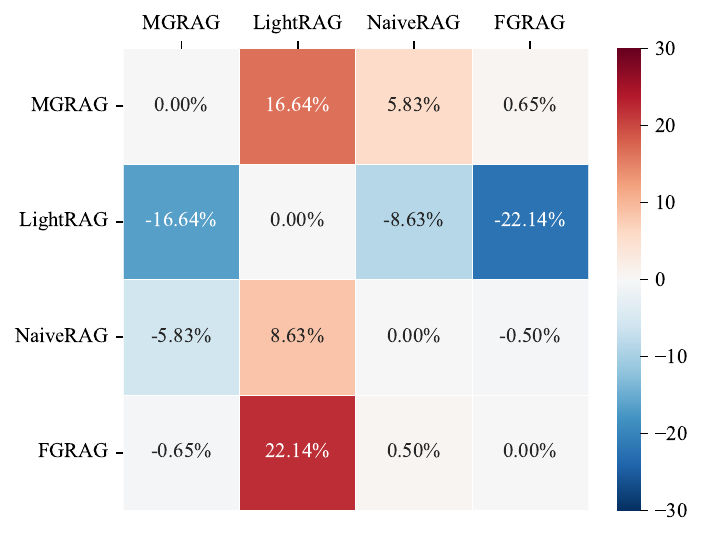}
        \caption{Node-level.}
        \label{fig:node}
        \vspace{8pt}
    \end{subfigure}
    \hfill
    \centering 
    \begin{subfigure}[t]{0.3\textwidth} 
        \centering
        \includegraphics[width=\linewidth]{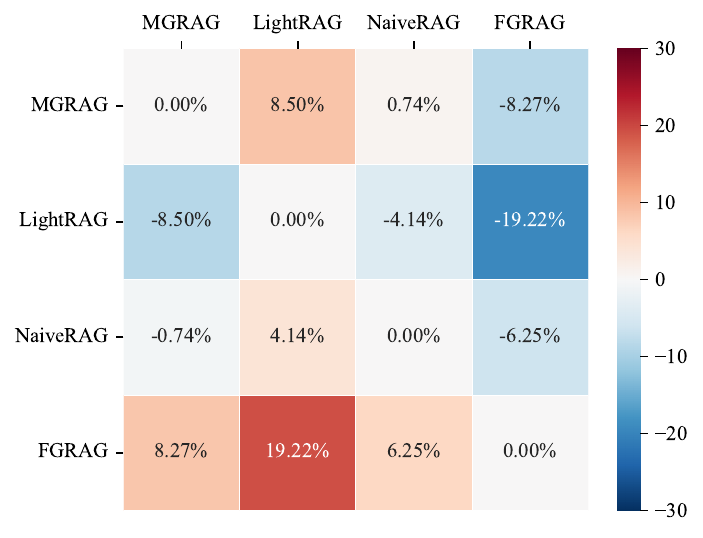} 
        \caption{Edge-level.} 
        \label{fig:edge} 
    \end{subfigure}
    \hfill
    \begin{subfigure}[t]{0.3\textwidth} 
        \centering
        \includegraphics[width=\linewidth]{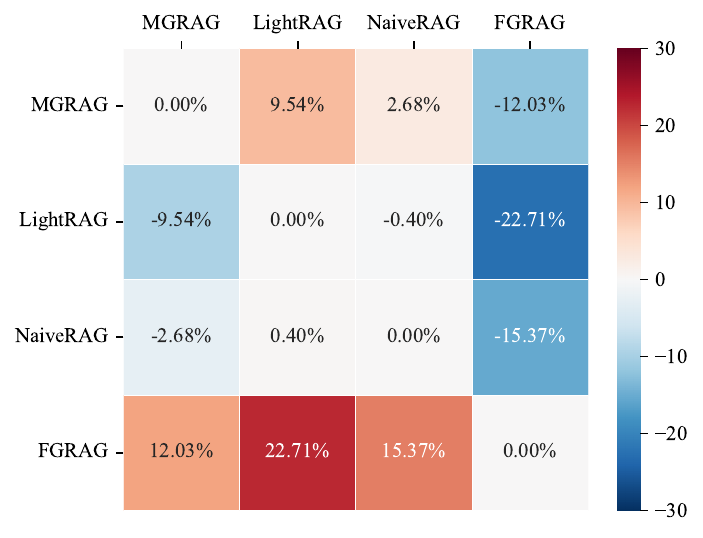} 
        \caption{Subgraph-level.} 
        \label{fig:subgraph} 
    \end{subfigure}
    \caption{The average win rate comparison across the four baseline model pairs evaluated on node-level, edge-level and subgraph-level questions under three datasets (i.e., Mix, Agriculture, and Music).} 
    \label{fig:level-qestins} 
\end{figure*}

\stitle{Detailed performance.}
In Figure \ref{fig:detail}, we look into how the methods perform in the detailed aspects of answer quality (i.e., comprehensiveness, relevance, empowerment, and directness). We only report the results on the Music dataset due to the page limit. We make the following observations.

First, the tie rates are much higher than the overall answer quality results in Figure~\ref{fig:overall}. In particular, the tie rates are above 50\% for many cases and seldom below 25\%. This suggests that the answers generated by different methods are likely to have similar quality in individual evaluation aspects, even if our evaluation decides that a method wins another for a question, the actual quality gap between their answers can be small.

Second, the performance of the methods for the detailed aspects reflects their designs. For instance, when comparing FGRAG and MGRAG, FGRAG excels in comprehensiveness, while MGRAG leads in directness. This is because FGRAG retrieves broader knowledge, which enhances comprehensiveness, while MGRAG directly leverages the relevant summaries for more targeted answers.

Third, when comparing NaiveRAG and GraphRAG, GraphRAG consistently leads in comprehensiveness, even LightRAG, with a lower win rate than NaiveRAG. This confirms that GraphRAG, by incorporating knowledge graphs, better captures cross-text content and generates more comprehensive answers than direct document retrieval.

\begin{mdframed}
\em \textbf{Findings:} Ties should be allowed for questions without perfect answer, as it do occurs. Different GraphRAG-specific designs directly affect the answer quality in specific aspects. The introduction of knowledge graphs can ultimately enhance the comprehensiveness of answers.
\end{mdframed}

\stitle{Performance across different question types.}
In Figure~\ref{fig:level-qestins}, we show the average win rates of different GraphRAG methods over different level questions to further explore the impact of various methods. We make the following observations.


First, our experimental results indicate that MGRAG achieves superior performance on node-level questions, which can be attributed to its ability to generate highly fine-grained community reports with minimal redundancy while maintaining rich information for a limited number of entities. In contrast, FGRAG’s overall performance follows a hierarchical ranking: subgraph level outperforms edge level, which in turn outperforms node level. This aligns with its original design objective of addressing complex problems that require a comprehensive understanding of the dataset. Analyzing FGRAG's retrieved information, we found its node-level summaries are less detailed, likely explaining its weaker performance here.




Second, LightRAG, another GraphRAG method, shows only small improvements on complex questions and little advantage over other approaches. This is likely because it includes too much information from one-hop expansions (adding noise) and lacks a good ranking system like FGRAG’s PageRank to highlight relevant details.
NaiveRAG performs the opposite of GraphRAG methods: it works much worse on subgraph-level questions than on node or edge levels. This fits its design, using fixed-length text segments for retrieval, which isn’t enough for complex subgraph questions.


\begin{mdframed}
\em \textbf{Findings:} GraphRAG’s design influences its performance across question types. The introduction of knowledge graphs enhance the model’s macro perspective, aiding broad-scope question answering.
\end{mdframed}

\stitle{Execution costs.} We also test the methods for their average number of consumed tokens and query time for answering a question, and Table~\ref{tab: tokentime} reports the results. The results are averaged over 100 questions based on the Mix dataset, and the query time is the duration from question submission to answer generation, which includes API request time. The results show that MGRAG has the highest token consumption and the longest query time. This is due to its need to retrieve and sort community summaries, which involves repeated LLM calls.
NaiveRAG is the most efficient in terms of token consumption, while FGRAG achieves the shortest query time. LightRAG ranks the second most in both metrics, and NaiveRAG’s performance is nearly on par with FGRAG in query time.

\begin{mdframed}
\em \textbf{Findings:} GraphRAG incurs relatively higher token overhead. Introducing re-ranking to reduce final retrieved information will be meaningful for improving efficiency.
\end{mdframed}

\renewcommand{\arraystretch}{1.3}
\begin{table}
  \caption{Average token consumption and query time }
  \begin{center}
  \label{tab: tokentime}
  \begin{tabular}{ccc}
    \toprule
      Method name & Token consumption & Query time (s)\\
    \midrule
    MGRAG & 326,200 & 25.84\\
    LightRAG & 19,650 & 18.27 \\
    NaiveRAG & \textbf{4,213} & 11.55 \\
    FGRAG & 16529 & \textbf{8.77}\\
    \bottomrule
  \end{tabular}
  \end{center}
\end{table}

%% file: sections/conclution.tex
\section{Conclusions and Future Directions}

By addressing limitations such as irrelevant questions and evaluation biases, we propose a novel framework based on graph-grounded question generation, alongside an unbiased evaluation procedure. We have experimentally demonstrated that our graph-text-grounded questions exhibit superior quality. Through the application of our unbiased evaluation framework to three GraphRAG methods and NaiveRAG, we demonstrate that their performance gains are more moderate than previously reported. While our framework may still have limitations, it underscores the importance of establishing a scientific and rigorous evaluation protocol to ensure the reliability and validity of future GraphRAG advancements.

%% file: sections/AI.tex
\section*{AI-Generated Content Acknowledgement}

The authors used an Al language model to assist with language polishing, solely to improve the clarity and fluency of the text. All technical content, analysis, experimental design, and conclusions were independently conceived and written by the authors.